\documentclass[11pt]{article}

\usepackage[final]{acl}

\usepackage{times}
\usepackage{latexsym}
\usepackage{graphicx}
\usepackage{booktabs}
\usepackage{multirow}
\usepackage{adjustbox}

\usepackage[most]{tcolorbox}
\tcbuselibrary{breakable,skins,listings}

\usepackage{caption}
\usepackage{needspace}
\usepackage{xcolor}
\usepackage{listings}

\usepackage{tabularx}

\usepackage{listings}
\usepackage[most]{tcolorbox}

\definecolor{promptblue}{RGB}{102,167,208}
\definecolor{promptgray}{gray}{0.4}

\definecolor{auditgreen}{RGB}{34,139,34}
\definecolor{auditlightgreen}{RGB}{248,255,248}
\definecolor{auditgray}{gray}{0.4}

\definecolor{extractionorange}{RGB}{217,95,2}
\definecolor{extractionlightorange}{RGB}{255,248,240}
\definecolor{extractiongray}{gray}{0.4}

\lstdefinestyle{extractionpromptstyle}{
  basicstyle=\ttfamily\scriptsize,
  breaklines=true,
  breakatwhitespace=false,
  columns=fullflexible,
  keepspaces=true,
  showstringspaces=false,
  xleftmargin=0pt,
  xrightmargin=0pt,
  aboveskip=0pt,
  belowskip=0pt
}

\lstdefinestyle{auditpromptstyle}{
  basicstyle=\ttfamily\scriptsize,
  breaklines=true,
  breakatwhitespace=false,
  columns=fullflexible,
  keepspaces=true,
  showstringspaces=false,
  xleftmargin=0pt,
  xrightmargin=0pt,
  aboveskip=0pt,
  belowskip=0pt
}

\lstdefinestyle{promptstyle}{
  basicstyle=\ttfamily\scriptsize,
  breaklines=true,
  breakatwhitespace=false,
  columns=fullflexible,
  keepspaces=true,
  showstringspaces=false,
  xleftmargin=0pt,
  xrightmargin=0pt,
  aboveskip=0pt,
  belowskip=0pt
}

\usepackage[T1]{fontenc}

\usepackage[utf8]{inputenc}

\usepackage{microtype}

\usepackage{inconsolata}

\usepackage{graphicx}
\usepackage{pifont}
\usepackage[normalem]{ulem}
\usepackage{marvosym}

\usepackage[table]{xcolor}

\usepackage{xcolor}  
\definecolor{cvprblue}{rgb}{0.21,0.49,0.74}  

\usepackage{graphicx}
\usepackage{wrapfig}
\usepackage{multirow}
\usepackage{booktabs}
\usepackage[table]{xcolor}
\usepackage{relsize}
\usepackage[most]{tcolorbox}
\usepackage{enumitem}
\usepackage{needspace}

\definecolor{mygreen}{RGB}{0,150,0}
\definecolor{myred}{RGB}{200,0,0}
\definecolor{lightblue}{RGB}{102, 167, 208}
\definecolor{lightgray}{RGB}{238, 238, 238}
\definecolor{lightgray}{gray}{0.9}
\definecolor{mygreen}{RGB}{0,150,0}
\definecolor{myred}{RGB}{200,0,0}
\def\BibTeX{{\rm B\kern-.05em{\sc i\kern-.025em b}\kern-.08em
    T\kern-.1667em\lower.7ex\hbox{E}\kern-.125emX}}

\title{PhotoCraft: Agentic Reasoning with Hierarchical Self-Evolving Memory for Deep Image Search}

\author{
\textbf{Kailin Lyu}$^{1 ,2,3,\dagger}$,
\textbf{Zhiqiang Yuan}$^{1,\dagger,\S}$,
\textbf{Jianwei He}$^{2}$,
\textbf{Qiwei Yan}$^{1}$,
\textbf{Xuanbo Su}$^{2}$,
\textbf{Nanxing Hu}$^{1}$
\\
\textbf{Yang Liu}$^{1}$,
\textbf{Ce Hao}$^{3}$,
\textbf{Shengqian Qin}$^{5}$,
\textbf{Lianyu Hu}$^{4,*}$,
\textbf{Jinchao Zhang}$^{1,*}$,
\textbf{Jie Zhou}$^{1}$
\\[0.4em]
$^{1}$Pattern Recognition Center, WeChat AI, Tencent Inc.
\\
$^{2}$Institute of Automation, Chinese Academy of Sciences
\\
$^{3}$Zhongguancun Academy
\\
$^{4}$Nanyang Technological University
\\
$^{5}$Shanghai Jiao Tong University
}

\begin{document}
\maketitle

\begingroup
\renewcommand{\thefootnote}{\fnsymbol{footnote}}
\footnotetext[2]{Equal contribution. $^{\S}$ Tech Lead. * Corresponding authors. $\ddagger$ This work was done during Kailin Lyu's internship at WeChat AI, Tencent Inc under guidance of Zhiqiang Yuan.}
\endgroup

\begin{abstract}
Deep Image Search requires multi-step reasoning over rich contextual cues, such as time, location, and event relations. However, most existing LLM-based agents are stateless and reactive, lacking persistent memory to maintain long-horizon context or transfer experience across tasks, which often leads to execution drift and experience isolation. To address these limitations, we propose PhotoCraft, a training-free, hierarchical memory system for photo-search agents. Inspired by human cognition, PhotoCraft equips MLLMs with working, episodic, and semantic memory, which are dynamically invoked during reasoning to preserve logical consistency and knowledge transferability throughout multi-step reasoning and answer generation. Extensive experiments on DISBench demonstrate that PhotoCraft consistently improves context-aware retrieval across diverse MLLM backbones, achieving gains of up to 18.5\% and effectively mitigating key bottlenecks in memoryless deep image search, offering a practical path toward reliable and generalizable multimodal search agents. Our code will be released.
\end{abstract}

\section{Introduction}
\label{Introduction}
\begin{figure}[!t]
    \centering
    \includegraphics[width=\linewidth]{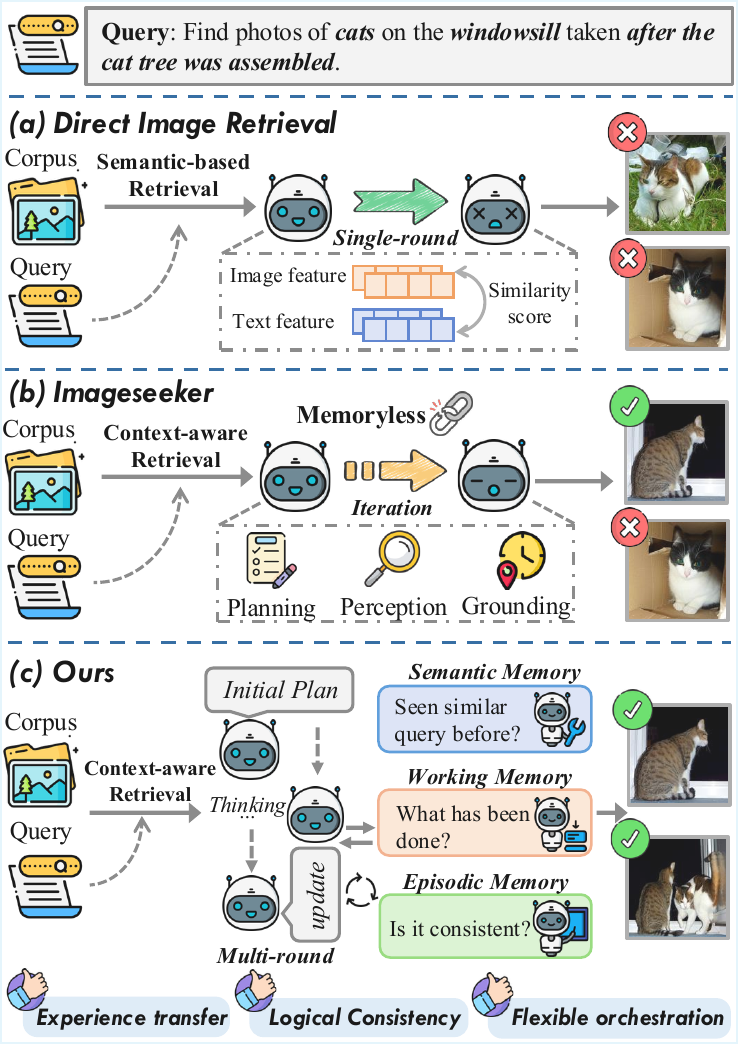}
    \caption{Paradigm shift from semantic matching to self-evolving, intent aware image retrieval. (a) Semantic retrieval depends on explicit query image overlap. (b) ImageSeeker performs iterative contextual reasoning, but its memoryless design degrades over multiple rounds. (c) PhotoCraft updates memory throughout multi step reasoning, enabling more reliable retrieval.}
    \label{fig:intro1}
\end{figure}

In real world personal photo retrieval, users rarely rely solely on precise visual descriptions. Instead, they retrieve images by reasoning over contextual cues such as time, location, and cross image associations. To capture this setting, Deep Image Search~\citep{deepimagesearch} reformulates image retrieval as an autonomous exploration task over personal visual histories. As shown in Figures~\ref{fig:intro1}, rather than independently matching a text query to each image based on semantic similarity~\citep{traditionalretrieval,traditionalretrieval2}, DeepImageSearch highlights context-aware retrieval, which requires models to: (1) integrate visual content, spatiotemporal metadata, and cross-event associations; (2) plan tool-assisted search trajectories conditioned on user intent; and (3) construct evidence chains to localize target images. Together, these characteristics make multi-step contextual reasoning an essential capability for MLLMs.

\begin{figure*}[!t] 
    \centering 
    \includegraphics[width=\textwidth]{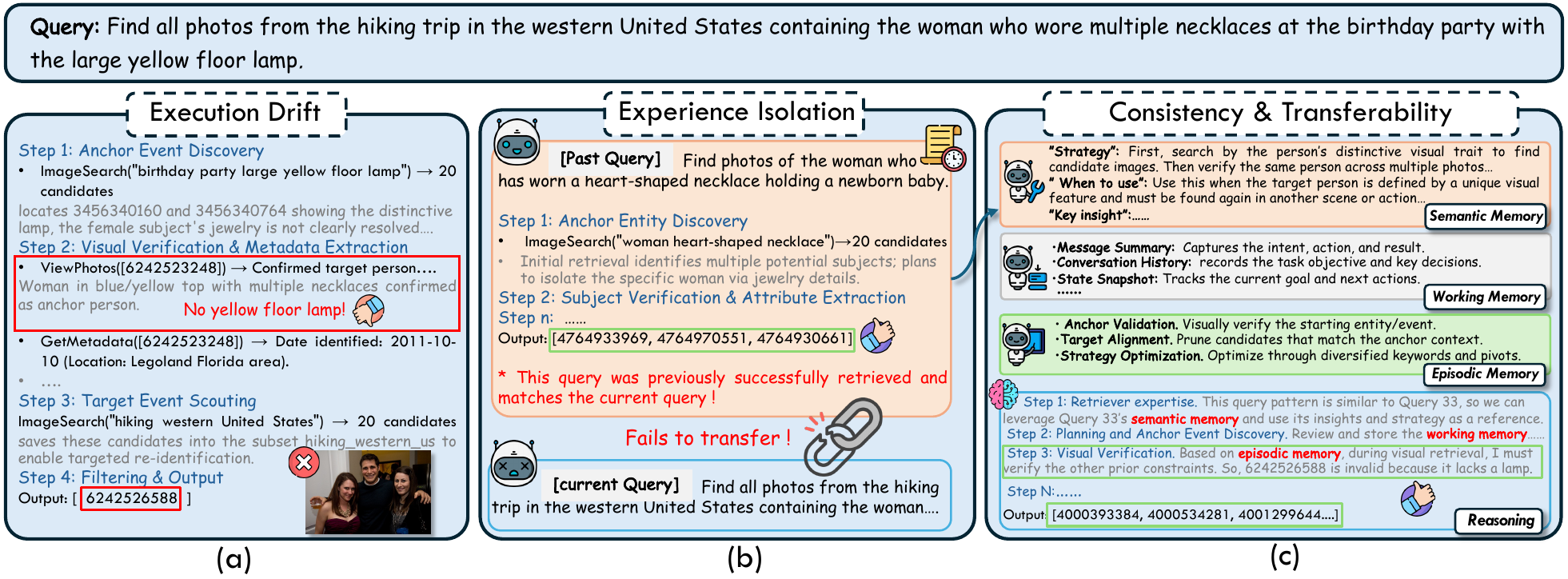}
    \caption{Two fundamental bottlenecks of memoryless search agents and PhotoCraft's hierarchical memory solution.} 
    \label{fig:intro2} 
\end{figure*}

To bridge the gap between passive retrieval and active exploration, existing search agent frameworks~\citep{deepimagesearch} employ MLLMs with pre-orchestrated tools and limited context compression mechanisms to enable multi step interaction to some extent. However, despite their strong reasoning capabilities, the stateless nature of MLLMs prevents them from maintaining coherent multi-step context within a single episode or accumulating knowledge across tasks~\citep{evolver,xskill}. This lack of memory introduces fundamental bottlenecks for search agents along two key dimensions: \ding{182} \textbf{\textit{Insufficient intra task reasoning consistency}}: As shown in Figure~\ref{fig:intro2}(a), without dynamic refocus guidance and effective memory compression, the agent may forget or misuse key evidence from earlier steps, causing cross-step constraint inconsistency that breaks the reasoning chain and leads to execution drift. \ding{183} \textbf{\textit{Limited flexibility in cross task orchestration}}: Constrained by a single path execution pattern, the agent must replan from scratch for each new query, making strategy transfer difficult. As shown in Figure~\ref{fig:intro2}(b), even after successfully solving structurally similar queries, the agent may still fail unexpectedly on a new task, resulting in experience isolation. In contrast, humans naturally rely on hierarchical memory systems to support dynamic interaction, continual adaptation, and sustained improvement in problem solving. This naturally raises a central question: \textit{How can we build a hierarchical memory system for agents that continuously enhances both intra task reasoning robustness and cross task orchestration flexibility?}

To overcome this problem, we resort to cognitive psychology, specifically the Atkinson-Shiffrin Memory Theory~\citep{atkinson}: \textit{\uline{Human cognition relies on three complementary memory systems. Working memory supports short term situational awareness, episodic memory links the current process to historical goals, and semantic memory enables abstract knowledge representation and generalization.}} While this cognitive theory reveals the essence of human cognition, it can be smoothly translated into an architectural principle of search agents. Working memory maintains recent reasoning outcomes and agent states for stable short-term reasoning. Episodic memory links the current process to target intent and step-level planning, enabling reflective self-correction during multi-step reasoning. Semantic memory distills successful patterns into reusable principles for current decisions and future reuse.



Based on such inspiration, we propose \textbf{PhotoCraft}, a training-free and continually evolving tri-stream reasoning framework that can be plugged into multimodal search agents. As shown in Figure~\ref{fig:intro2}(c), PhotoCraft consists of three complementary memory systems: working memory, episodic memory, and semantic memory. We first design a dynamic buffer that maintains goal relevant constraint summaries, conversation history, and state snapshots, thereby realizing working memory for the current task state and intermediate context. During multi-step reasoning, the agent further collaborates with multimodal tools to localize evidence, interpret outcomes, and perform reflective correction, progressively forming episodic memory that enhances MLLM-based task understanding and reasoning consistency. Building on the above, we introduce a skill abstraction mechanism that distills successful trajectories into reusable structured strategies. For each new query, PhotoCraft dynamically retrieves relevant skills and injects lightweight directional guidance into the current context to simulate semantic memory, enabling robust strategy transfer while preserving autonomous exploration and to evolve from a one-shot solver into a continually improving self-evolving system. The complementary integration of different memory modules further improves multi-step reasoning consistency and cross-task transferability. Extensive experiments on DISBench~\citep{deepimagesearch} show that our method consistently improves context aware image retrieval across multiple base models and effectively alleviates the limitations of memoryless reasoning. Our contributions are listed as follows:

\begin{itemize}[leftmargin=*, itemsep=0pt, parsep=0pt, topsep=2pt, partopsep=0pt]
    \item We identify Execution Drift and Experience Isolation as two critical bottlenecks in Deep Image Search, and propose PhotoCraft, a training-free and self-evolving framework.
    \item We propose a hierarchical memory system for search agents, with distinct functions and mechanisms inspired by cognitive psychology. This design improves both situational awareness and reasoning consistency.
    \item Extensive quantitative and qualitative experiments demonstrate that the framework significantly improves both contextual reasoning and cross task generalization.
\end{itemize}

\section{Related Works}
\label{Related Works}


\noindent \textbf{Image Retrieval.} Image retrieval aims to identify relevant visual content from large-scale image collections based on user queries~\citep{traditionalretrieval,imageretrieversurvey2}. Recent vision-language pretrained models align images and texts in a shared embedding space and retrieve candidates through similarity matching, enabling scalable open-vocabulary visual search~\citep{vlmimageretriever1,vlmimageretriever2}. Extensions such as composed image retrieval~\citep{composedretr,composedsurvey} and reasoning-enhanced retrieval~\citep{thinktheret,reaoningret} further incorporate textual modifications, query rewriting, or external knowledge to capture complex user intents. However, most methods still score each candidate independently, making them effective for visually explicit targets but insufficient for context-dependent retrieval. Therefore, Deng et al.~\citep{deepimagesearch} attempt to address this issue through the Deep Image Search paradigm, which formulates retrieval as agentic corpus exploration. However, its reliance on stateless MLLM interactions weakens long-horizon reasoning and limits generalization. In contrast, PhotoCraft introduces a memory-aware agentic retrieval framework that maintains intermediate evidence, models cross-image dependencies, and enables stable reasoning over visual histories.


\noindent \textbf{Agentic Reasoning for Multimodal Tasks.} Agent systems built on LLMs and MLLMs have recently emerged as an important paradigm for tasks requiring sequential decision making and adaptive information seeking~\citep{multiagentsurvey,agentsurvey}. Rather than relying on a single reasoning pass, they decompose goals, invoke tools, update memory~\citep{tree}, and refine reasoning from new evidence. This paradigm has been applied to multimodal question answering~\citep{qaavis,searchr1}, embodied intelligence~\citep{embodiedeval,roboagent}, and long context video understanding~\citep{vgent,videoarm}. ImageSeeker~\citep{deepimagesearch} first explores agentic image retrieval with tools such as location grounding and visual inspection, and evaluates it on DISBench. However, it relies on pre-orchestrated iterative contextual reasoning, while its memoryless design becomes less effective over multiple rounds. In contrast, inspired by cognitive psychology~\citep{atkinson}, we propose a hierarchical memory system for deep image retrieval agents, enabling distinct memory modules to enhance reasoning over corpus retrieval.

\section{Method}
\label{Method}


In deep image search, a single query often requires dozens of tool calls and reasoning steps. However, as discussed in Section~\ref{Introduction}, MLLM-based stateless agents suffer from a memory bottleneck that induces execution drift within tasks and hinders experience transfer across tasks. To address this, we propose PhotoCraft, a search agent equipped with a hierarchical self-evolving memory system. Inspired by the hierarchical organization of human cognition, the system is structured along the abstraction spectrum of perception, reflection, and generalization into three complementary components: \textit{working memory, episodic memory, and semantic memory}. The MLLM serves as the core controller, autonomously orchestrating a suite of multimodal tools (e.g., visual verification) within an iterative loop while continuously maintaining the memory system to progressively narrow the search space and localize the target images. An overview of PhotoCraft is shown in Figure~\ref{fig:pipline}, with tool details provided in the \textbf{Appendix}~\ref{Tool Description}.

\begin{figure*}[!t] 
    \centering 
    \includegraphics[width=\textwidth]{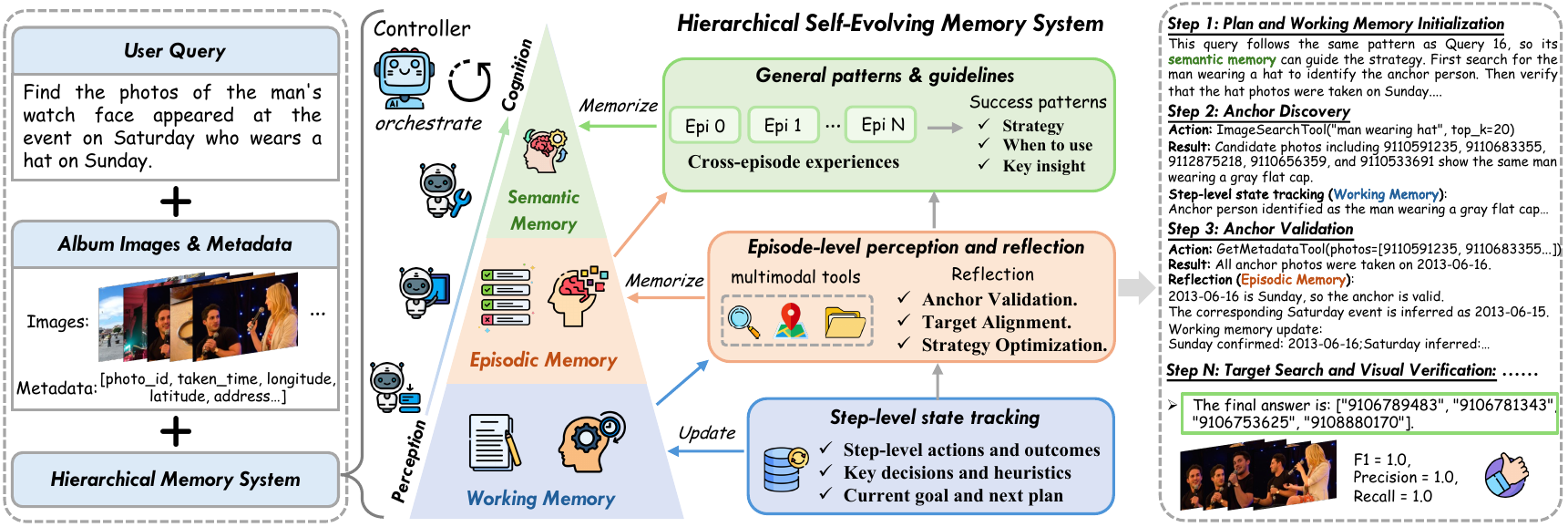}
    \vspace{-6mm}
    \caption{Overview of the PhotoCraft and an example of its reasoning pipeline.} 
    \label{fig:pipline} 
    \vspace{-4mm}
\end{figure*}

\subsection{Task Formulation}
\label{Task Formulation}

We formalize Deep Image Search as a context-aware set retrieval task.
Given a user's visual history corpus \(C = \{I_1, I_2, \ldots, I_N\}\), each image \(I_i = (v_i, m_i)\) consists of visual content \(v_i\) and metadata \(m_i\), where \(m_i\) includes a timestamp \(\tau_i\) and geographic coordinates \(g_i\). Upon receiving a natural language query \(Q\), the system is required to predict a target subset \(R \subseteq C\) that contains all images satisfying the semantic intent of \(Q\). Unlike
conventional retrieval paradigms that score each image independently,
Deep Image Search requires modeling the joint posterior \(P(R \mid Q, C)\),
where the relevance of an individual image may depend on contextual cues
provided by other images in the corpus. In this task, queries are provided in text form, while visual references are uniformly converted into
textual descriptions. This design requires the model to first identify
visual anchors within \(C\) and then perform spatiotemporal reasoning based
on these anchors, making the task substantially more challenging than direct visual matching.

\subsection{Hierarchical Self-Evolving Memory}
\label{Hierarchical Self-Evolving Memory}



\noindent\raisebox{-.3\baselineskip}{\includegraphics[height=1.2\baselineskip]{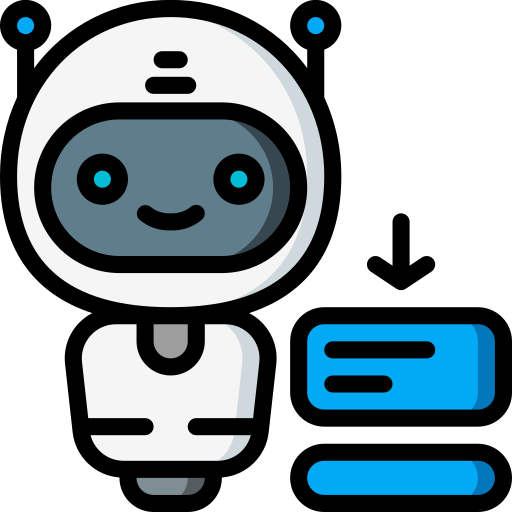}}~\textbf{Working Memory.} Inspired by human-like working memory and its selective maintenance of task relevant information, we argue that meaningful short term memory should not be measured by a fixed retained window, but by its ability to preserve valid state information throughout ongoing reasoning. Therefore, unlike conventional free form compression within a fixed context window~\citep{deepimagesearch}, we introduce fidelity guarded working memory $\mathcal{M}_{\mathrm{work}}$, a lightweight memory compression mechanism for preserving short-term reasoning state. Specially, When the context approaches the window limit, the memory module \(f_{\mathrm{mem}}\) compresses the message history \(\mathcal{H}_c=\{m_1,\ldots,m_T\}\) into three complementary summaries:
\begin{equation}
S=f_{\mathrm{mem}}(\mathcal{H}_c,Q)
= S_{\mathrm{digest}} \oplus S_{\mathrm{session}} \oplus S_{\mathrm{working}},
\label{eq1}
\end{equation}
Here, \(S_{\mathrm{digest}}\) preserves step level actions and outcomes for traceability, \(S_{\mathrm{session}}\) records key decisions and reusable heuristics for strategic continuity, and \(S_{\mathrm{working}}\) stores the current subgoal and next plan for seamless continuation. To prevent metadata loss in compressed memory, we further apply a deterministic fidelity validator \(\psi\). After generating summary \(S\), the fidelity validator \(\psi\) extracts all tool messages containing metadata \(\mathcal{M}_{\mathrm{meta}} \subseteq \mathcal{H}_c\) from the original history and checks whether critical fields, including photo IDs, temporal constraints, and spatial filters, are preserved in \(S\). The metadata loss rate is defined as:
\begin{equation}
\mathcal{L} = 1 - \frac{\sum_{m \in \mathcal{M}_{\mathrm{meta}}} \mathrm{Preserved}(m,S)}{|\mathcal{M}_{\mathrm{meta}}|},
\end{equation}
where \(\mathrm{Preserved}(m,S)\) is an indicator that equals \(1\) only when all constraints in message \(m\) are retained in \(S\), and \(0\) otherwise.
If \(\mathcal{L}>\tau\), with \(\tau=0.5\) by default, the compression is deemed unreliable, and the most recent metadata critical messages are retained in their original form. Otherwise, standard compression is performed.


\noindent\raisebox{-.3\baselineskip}{\includegraphics[height=1.2\baselineskip]{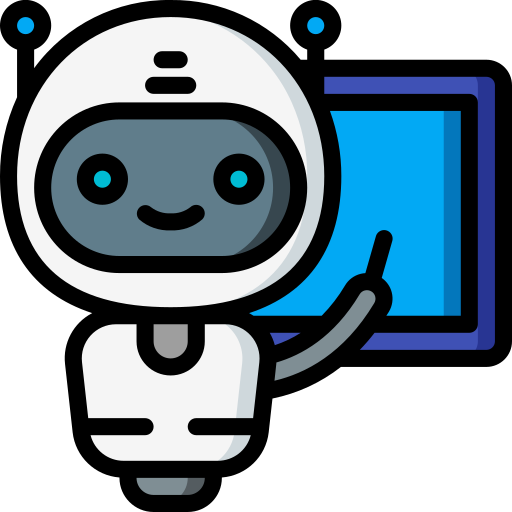}}~\textbf{Episodic Memory.} Inspired by the human capacity to reconstruct events within a specific spatiotemporal context, we introduce episodic memory, denoted as \(\mathcal{M}_{\mathrm{epi}}\). This module performs reflective self-correction within a single episode through an explicit posthoc verification stage after the agent generates a candidate answer, rather than allowing the MLLM to produce the final output through a single forward pass. Formally, given a candidate answer \(\hat{R}\), the episodic memory module \(f_{\mathrm{epi}}\) is conditioned on the query \(Q\), the compressed history \(\mathcal{H}_c\), and a visual reload \(V(\hat{R})\) of the photos referenced by \(\hat{R}\), and produces a structured audit signal:
\begin{equation}
\begin{aligned}
E &= f_{\mathrm{epi}}\bigl(Q, H_c, V(\hat{R})\bigr),
\end{aligned}
\label{eq:episodic-memory}
\end{equation}
The audit signal \(E\) decomposes into five components, \{\textsc{Anchor}, \textsc{Constraint}, \textsc{Target}, \textsc{Completeness}, and \textsc{Decision}\}, which jointly perform a comprehensive verification across complementary dimensions,  with detailed descriptions provided in the \textbf{Appendix}~\ref{Prompts for Episodic Audit Signals}. When the audit judges the trajectory as inconsistent, \(\mathcal{M}_{\mathrm{epi}}\) appends the verdict to the working memory as structured feedback and prompts the agent to launch a new retrieval round with differentiated keywords, thereby explicitly triggering supplementary evidence gathering through context re-orchestration. In this way, \(\mathcal{M}_{\mathrm{epi}}\) empowers the agent to reconstruct the reasoning trajectory within the current spatiotemporal context and perform reflective self correction, mitigating execution drift and improving logical consistency in complex multi hop queries.

\noindent\raisebox{-.3\baselineskip}{\includegraphics[height=1.2\baselineskip]{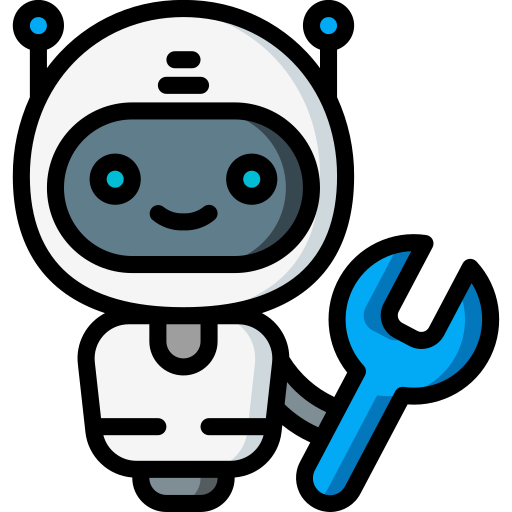}}~\textbf{Semantic Memory.} Inspired by the decontextualized abstraction of human semantic memory, the guiding intuition of semantic memory $\mathcal{M}_{\mathrm{sem}}$ is to avoid replanning from scratch for each query by distilling task-decoupled strategy prototypes from successful trajectories and injecting them as directional heuristics when appropriate. Accordingly, $\mathcal{M}_{\mathrm{sem}}$ consists of a skill distiller and an evolving skill library, operating under the following pipeline: 

\noindent \textit{\textbf{(1) Skill Distillation.}} Upon the completion of each episode, we regard it as a successful experience if its final F1 score exceeds the confidence threshold $\theta_{\mathrm{conf}}$.
The distiller then feeds the entire trajectory $\mathcal{T} = \{(R_t, A_t, P_t, O_t)\}_{t=1}^{T}$ into the skill module to produce a structured skill entry $m_{\mathrm{sem}}=f_{\mathrm{skill}}(\mathcal{T})$, which contains \{\textsc{task\_pattern}, \textsc{strategy}, \textsc{key\_insight}, \textsc{applicable\_when}, and \textsc{confidence}\}. Detailed field definitions are provided in the \textbf{Appendix}~\ref{Prompts for Semantic Skill Entries}, with an illustrative example shown in Figure~\ref{sem_example}. Distillation is further constrained by an explicit de-entitification requirement that strips episode-specific surface entities and retains only transferable query structures and solution strategies, so that each $m_{\mathrm{sem}}$ acts as a genuine abstract pattern rather than a memorized instance. 

\noindent \textit{\textbf{(2) Skill Consolidation.}} We employ a lightweight sentence-transformer encoder $\phi(\cdot)$ to map the source query of each skill into a semantic vector~\citep{sentence}, upon which we perform a similarity-based merge-and-grow strategy. For a newly distilled skill $m_{\mathrm{sem}}^{\mathrm{new}}$ with source query $q^{\mathrm{new}}$, if there exists an incumbent skill $m_{\mathrm{sem}}^{j} \in \mathcal{M}_{\mathrm{sem}}$ with source query $q^{j}$ such that 
\begin{equation} 
\cos\!\left(\phi(q^{\mathrm{new}}),\, \phi(q^{j})\right) > 0.8,
\label{eq:skill-merge} 
\end{equation}
$m_{\mathrm{sem}}^{\mathrm{new}}$ is merged into $m_{\mathrm{sem}}^{j}$; otherwise it is admitted as an independent entry, i.e., $\mathcal{M}_{\mathrm{sem}} \leftarrow \mathcal{M}_{\mathrm{sem}} \cup \{m_{\mathrm{sem}}^{\mathrm{new}}\}$. This mechanism prevents redundant bloat of the library while allowing recurrent effective patterns to be naturally reinforced, yielding a more compact and higher-quality $\mathcal{M}_{\mathrm{sem}}$.

\noindent \textit{\textbf{(3) Skill Retrieval and Injection.}} Given a new query \(Q\), we compute cosine similarities between \(\phi(Q)\) and the source query embeddings of all entries in \(\mathcal{M}_{\mathrm{sem}}\). The most similar entry is retrieved only when its score exceeds the retrieval threshold \(\theta_{\mathrm{retr}}\), after which it is prepended to the user message as a directional hint. Notably, this design preserves the agent's autonomy in deciding whether to use the hint, while reducing redundant exploration and repeated failures when prior experience is well aligned with the current task.


\begin{table*}[!t]
\centering
\setlength{\tabcolsep}{2pt}
\renewcommand{\arraystretch}{1.2}
\caption{Performance comparison (\%) between PhotoCraft and the ImageSeeker baseline on DISBench. Superscripts denote absolute improvements over the baseline reported in DeepImageSearch~\citep{deepimagesearch}. \textbf{Bold} and \underline{underline} values indicate the best and second best results, respectively.}
\vspace{-2mm}
\label{tab1}
\resizebox{\textwidth}{!}{
\begin{tabular}{lcccccccccccc}
\toprule
\multirow{3}{*}{\textbf{Base Model}}
& \multicolumn{6}{c}{\textbf{Qwen3-VL-Embedding-2B}}
& \multicolumn{6}{c}{\textbf{Qwen3-VL-Embedding-8B}} \\
\cmidrule(lr){2-7} \cmidrule(lr){8-13}
& \multicolumn{2}{c}{\textbf{Intra-Event}}
& \multicolumn{2}{c}{\textbf{Inter-Event}}
& \multicolumn{2}{c}{\textbf{Overall}}
& \multicolumn{2}{c}{\textbf{Intra-Event}}
& \multicolumn{2}{c}{\textbf{Inter-Event}}
& \multicolumn{2}{c}{\textbf{Overall}} \\
\cmidrule(lr){2-3} \cmidrule(lr){4-5} \cmidrule(lr){6-7}
\cmidrule(lr){8-9} \cmidrule(lr){10-11} \cmidrule(lr){12-13}
& \textbf{EM} & \textbf{F1}
& \textbf{EM} & \textbf{F1}
& \textbf{EM} & \textbf{F1}
& \textbf{EM} & \textbf{F1}
& \textbf{EM} & \textbf{F1}
& \textbf{EM} & \textbf{F1} \\
\specialrule{0.5pt}{0pt}{0pt}
\multicolumn{13}{>{\columncolor{lightgray}}c}{\textit{\textbf{Closed-Source Models}}} \\

GPT-4o
& 10.5$^{\textcolor{mygreen}{+5.2}}$
& 25.6$^{\textcolor{mygreen}{+6.0}}$
& 13.8$^{\textcolor{mygreen}{+4.6}}$
& 26.0$^{\textcolor{mygreen}{+1.5}}$
& 12.3$^{\textcolor{mygreen}{+4.9}}$
& 25.8$^{\textcolor{mygreen}{+3.6}}$
& 7.0$^{\textcolor{mygreen}{+1.7}}$
& 21.3$^{\textcolor{mygreen}{+4.2}}$
& 7.7$^{\textcolor{mygreen}{+1.5}}$
& 27.2$^{\textcolor{mygreen}{+1.3}}$
& 7.4$^{\textcolor{mygreen}{+1.7}}$
& 24.4$^{\textcolor{mygreen}{+2.6}}$ \\

GPT-5.2
& 19.3$^{\textcolor{mygreen}{+8.8}}$
& 44.6$^{\textcolor{mygreen}{+6.6}}$
& 15.4$^{\textcolor{mygreen}{+3.1}}$
& 36.9$^{\textcolor{mygreen}{+4.3}}$
& 17.2$^{\textcolor{mygreen}{+5.7}}$
& 40.5$^{\textcolor{mygreen}{+5.4}}$
& 26.3$^{\textcolor{mygreen}{+7.0}}$
& 42.8$^{\textcolor{mygreen}{+9.9}}$
& 10.8$^{\textcolor{mygreen}{+3.1}}$
& 29.5$^{\textcolor{mygreen}{+2.1}}$
& 18.0$^{\textcolor{mygreen}{+4.9}}$
& 35.7$^{\textcolor{mygreen}{+5.7}}$ \\

Gemini-3-Flash-Preview
& 19.3$^{\textcolor{mygreen}{+3.5}}$
& 40.7$^{\textcolor{mygreen}{+1.4}}$
& 9.2$^{\textcolor{mygreen}{+1.5}}$
& 30.7$^{\textcolor{mygreen}{+1.5}}$
& 13.9$^{\textcolor{mygreen}{+2.4}}$
& 35.4$^{\textcolor{mygreen}{+1.5}}$
& 17.5$^{\textcolor{mygreen}{+1.7}}$
& 43.3$^{\textcolor{mygreen}{+0.9}}$
& 9.2$^{\textcolor{mygreen}{+0.0}}$
& 32.4$^{\textcolor{mygreen}{+0.5}}$
& 13.1$^{\textcolor{mygreen}{+0.8}}$
& 37.5$^{\textcolor{mygreen}{+0.7}}$ \\

Claude-Sonnet-4.5
& \underline{26.3}$^{\textcolor{mygreen}{+3.5}}$
& \underline{47.3}$^{\textcolor{mygreen}{+3.3}}$
& \underline{18.5}$^{\textcolor{mygreen}{+6.2}}$
& \underline{38.0}$^{\textcolor{mygreen}{+2.6}}$
& \underline{22.1}$^{\textcolor{mygreen}{+4.9}}$
& \underline{42.3}$^{\textcolor{mygreen}{+2.9}}$
& \underline{29.8}$^{\textcolor{mygreen}{+1.7}}$
& \underline{51.9}$^{\textcolor{mygreen}{+3.4}}$
& \underline{18.5}$^{\textcolor{mygreen}{+1.6}}$
& \underline{42.7}$^{\textcolor{mygreen}{+3.1}}$
& \underline{23.8}$^{\textcolor{mygreen}{+1.7}}$
& \underline{47.0}$^{\textcolor{mygreen}{+3.2}}$ \\

Claude-Opus-4.5
& \textbf{38.6}$^{\textcolor{mygreen}{+3.5}}$
& \textbf{58.7}$^{\textcolor{mygreen}{+0.8}}$
& \textbf{35.4}$^{\textcolor{mygreen}{+6.2}}$
& \textbf{57.0}$^{\textcolor{mygreen}{+3.6}}$
& \textbf{36.9}$^{\textcolor{mygreen}{+4.9}}$
& \textbf{57.8}$^{\textcolor{mygreen}{+2.3}}$
& \textbf{40.4}$^{\textcolor{mygreen}{+5.3}}$
& \textbf{63.9}$^{\textcolor{mygreen}{+3.9}}$
& \textbf{29.2}$^{\textcolor{mygreen}{+6.1}}$
& \textbf{53.7}$^{\textcolor{mygreen}{+3.0}}$
& \textbf{34.4}$^{\textcolor{mygreen}{+5.7}}$
& \textbf{58.5}$^{\textcolor{mygreen}{+3.5}}$ \\

\specialrule{0.5pt}{0pt}{0pt}
\multicolumn{13}{>{\columncolor{lightblue!25}}c}{\textit{\textbf{Open-Source Models}}} \\

Qwen3-VL-235B-A22B-Thinking
& 14.0$^{\textcolor{mygreen}{+3.5}}$
& 25.6$^{\textcolor{mygreen}{+2.4}}$
& 12.3$^{\textcolor{mygreen}{+3.1}}$
& 24.5$^{\textcolor{mygreen}{+1.9}}$
& 13.1$^{\textcolor{mygreen}{+3.3}}$
& 25.0$^{\textcolor{mygreen}{+2.2}}$
& 15.8$^{\textcolor{mygreen}{+3.5}}$
& 25.6$^{\textcolor{mygreen}{+1.9}}$
& 9.2$^{\textcolor{mygreen}{+4.6}}$
& 21.6$^{\textcolor{mygreen}{+4.0}}$
& 12.3$^{\textcolor{mygreen}{+4.1}}$
& 23.5$^{\textcolor{mygreen}{+3.1}}$ \\

Qwen3-VL-235B-A22B-Instruct
& \underline{19.3}$^{\textcolor{mygreen}{+7.0}}$
& 28.6$^{\textcolor{mygreen}{+2.5}}$
& \underline{13.8}$^{\textcolor{mygreen}{+7.6}}$
& 26.4$^{\textcolor{mygreen}{+1.8}}$
& \underline{16.4}$^{\textcolor{mygreen}{+7.4}}$
& 27.4$^{\textcolor{mygreen}{+2.1}}$
& \underline{17.5}$^{\textcolor{mygreen}{+0.0}}$
& 32.6$^{\textcolor{mygreen}{+0.5}}$
& \underline{15.4}$^{\textcolor{mygreen}{+9.2}}$
& \underline{30.0}$^{\textcolor{mygreen}{+7.1}}$
& \underline{16.4}$^{\textcolor{mygreen}{+4.9}}$
& 31.2$^{\textcolor{mygreen}{+4.0}}$ \\

Qwen3-VL-32B-Instruct
& \textbf{26.3}$^{\textcolor{mygreen}{+10.5}}$
& \underline{36.3}$^{\textcolor{mygreen}{+4.3}}$
& \underline{13.8}$^{\textcolor{mygreen}{+7.6}}$
& \underline{31.2}$^{\textcolor{mygreen}{+11.6}}$
& \textbf{19.7}$^{\textcolor{mygreen}{+9.0}}$
& \underline{33.6}$^{\textcolor{mygreen}{+8.2}}$
& \textbf{21.1}$^{\textcolor{mygreen}{+7.1}}$
& \textbf{37.3}$^{\textcolor{mygreen}{+10.2}}$
& \textbf{21.5}$^{\textcolor{mygreen}{+18.4}}$
& \textbf{34.0}$^{\textcolor{mygreen}{+18.5}}$
& \textbf{21.3}$^{\textcolor{mygreen}{+13.1}}$
& \textbf{35.5}$^{\textcolor{mygreen}{+14.6}}$ \\

GLM-4.6V
& 17.5$^{\textcolor{mygreen}{+3.5}}$
& \textbf{38.5}$^{\textcolor{mygreen}{+4.3}}$
& \textbf{15.4}$^{\textcolor{mygreen}{+4.6}}$
& \textbf{35.3}$^{\textcolor{mygreen}{+8.3}}$
& \underline{16.4}$^{\textcolor{mygreen}{+4.1}}$
& \textbf{36.8}$^{\textcolor{mygreen}{+6.4}}$
& 12.3$^{\textcolor{mygreen}{+1.8}}$
& \underline{34.7}$^{\textcolor{mygreen}{+2.6}}$
& 9.2$^{\textcolor{mygreen}{+3.0}}$
& 28.3$^{\textcolor{mygreen}{+7.5}}$
& 10.7$^{\textcolor{mygreen}{+2.5}}$
& \underline{31.3}$^{\textcolor{mygreen}{+5.2}}$ \\

\bottomrule
\end{tabular}
}
\vspace{-3mm}
\end{table*}

\subsection{Agentic Reasoning Controller}
\label{Agentic Reasoning Controller}

To leverage the reasoning and planning capacity of MLLMs, we avoid rigid workflows and predefined tool use rules, and use the MLLM as the core controller. It operates autonomously in an loop, akin to ReAct style interaction~\citep{react}, supported by the hierarchical self evolving memory in Section~\ref{Hierarchical Self-Evolving Memory} as a persistent memory substrate. Specially, At each step, the controller invokes a selected tool with parameters $P_t$ and obtains an observation $O_t$. 
The observation is immediately written into $\mathcal{M}_{\mathrm{work}}$, where goal-relevant constraints, intermediate evidence, and state snapshots are updated and compressed when necessary to preserve a compact task context. 
Throughout the episode, $\mathcal{M}_{\mathrm{epi}}$ accumulates the step-level reasoning trajectory, including tool actions, observations, and intermediate decisions. 
Before the controller returns the final answer, $\mathcal{M}_{\mathrm{epi}}$ conducts a reflective audit over the accumulated trajectory. 
If inconsistencies or missing evidence are detected, the controller initiates an additional retrieval round with differentiated keywords until the audit passes or the step budget $N$ is reached. 
After the episode is completed, $\mathcal{M}_{\mathrm{sem}}$ evaluates the final trajectory based on the task outcome and distills successful patterns into reusable skill entries for future queries. This process progressively upgrades a stateless MLLM into a continually self-improving system without modifying model parameters.

\section{Experiment}
\label{Experiment}

\subsection{Experimental Settings}
\label{Experimental Settings}

\textbf{Dataset.} We conduct experiments on the DISBench~\citep{deepimagesearch} benchmark, which is constructed from YFCC100M~\citep{yfcc100m} and comprises 109,467 photos from 57 users, together with 122 queries rigorously selected from 2,000 candidates. Queries are categorized into two types: intra-event and inter-event. Unlike conventional text-to-image retrieval benchmarks, DISBench requires models to first locate visual anchors and then perform spatiotemporal reasoning, rather than relying on direct visual matching. Detailed descriptions are provided in the \textbf{Appendix}~\ref{Details of DISBench}.

\noindent \textbf{Implementation Details.} We evaluate the proposed PhotoCraft against the state-of-the-art ImageSeeker framework~\citep{deepimagesearch} under identical tool interfaces and memory budgets. Both frameworks are instantiated with a broad range of leading MLLMs serving as the backbone agent, including proprietary models: GPT-4o~\citep{gpt4o}, GPT-5.2, Gemini-3-Flash-Preview, Claude-Sonnet-4.5~\citep{sonnet45}, and Claude-Opus-4.5~\citep{opus45}; as well as open-source models: Qwen3-VL-235B-A22B-Thinking, Qwen3-VL-235B-A22B-Instruct, Qwen3-VL-32B~\citep{qwen3}, and GLM-4.6V~\citep{glm}. For the semantic memory module, we set the skill-distillation confidence threshold to $\theta_{\mathrm{conf}}$ = 0.85 and the retrieval-similarity threshold to $\theta_{\mathrm{retr}}$ = 0.65. We adopt Exact Match (EM) and F1 as evaluation metrics. Detailed tool descriptions and configuration settings are provided in the \textbf{Appendix}~\ref{Agent Implementation Details}.

\subsection{Main Results}
\label{Main Results}
Table~\ref{tab1} reports the main results of PhotoCraft on DISBench across multiple MLLMs. PhotoCraft consistently achieves the best performance across all metrics, outperforming the ImageSeeker~\citep{deepimagesearch} baseline with both proprietary and open source models. In particular, Qwen3-VL-32B-Instruct with PhotoCraft and Qwen3-VL-Embedding-2B substantially improves Overall EM and F1 by \textbf{9.0\%} and \textbf{8.2\%}, respectively, demonstrating its effectiveness in enhancing long horizon agentic reasoning. Notably, PhotoCraft also shows stronger stability. For example, ImageSeeker with Qwen3-VL-32B-Instruct performs well with the 2B embedding model but degrades sharply with the 8B embedding model, with EM and F1 falling to 3.1 and 15.5 on the Inter Event subset. In contrast, PhotoCraft remains robust across embedding scales, indicating that retrieval quality is not the primary bottleneck in Deep Image Search. Instead, the key challenge is reasoning over retrieved evidence, which PhotoCraft addresses through its hierarchical memory System. Moreover, Figure~\ref{reasoning_example} further provides qualitative evidence of PhotoCraft’s reasoning advantage over the baseline. These results collectively highlight the advantages of our framework and establish it as a practical and generalizable solution for Deep Image Search.

\subsection{Ablation Studies}
\label{Ablation Studies}
To analyze the contributions of different memory components in hierarchical self-evolving memory, we conduct a systematic ablation study on DISBench using Qwen3-VL-32B-Instruct and Qwen3-VL-Embedding-2B~\citep{qwen3}, as shown in Table~\ref{tab:ablation_memory}. The results show that each component addresses different challenges in agentic reasoning, and their integration is critical to the effectiveness of the framework. We analyze how each memory unit shapes distinct behavioral patterns as follow.

\begin{table}[!t]
\centering
\setlength{\tabcolsep}{2pt}
\renewcommand{\arraystretch}{1.2}
\small
\caption{Ablation study of different memory components on DISBench~\citep{deepimagesearch}. We report F1 scores. The full model achieves the best performance.}
\vspace{-2mm}
\label{tab:ablation_memory}
\begin{tabularx}{\columnwidth}{@{}Xccc@{}}
\toprule
\textbf{Model} & \textbf{Intra-event} & \textbf{Inter-event} & \textbf{Overall} \\
\midrule
\textbf{full model (PhotoCraft)} & \textbf{36.3} & \textbf{31.2} & \textbf{33.6} \\
\quad w/o Working Memory & 30.4 & 27.5 & 28.9 \\
\quad w/o Episodic Memory & 34.3 & 24.8 & 29.2 \\
\quad w/o Semantic Memory & 33.0 & 26.4 & 29.6 \\
\bottomrule
\end{tabularx}
\vspace{-5mm}
\end{table}

\begin{figure}[!t]
    \centering
    \includegraphics[width=\linewidth]{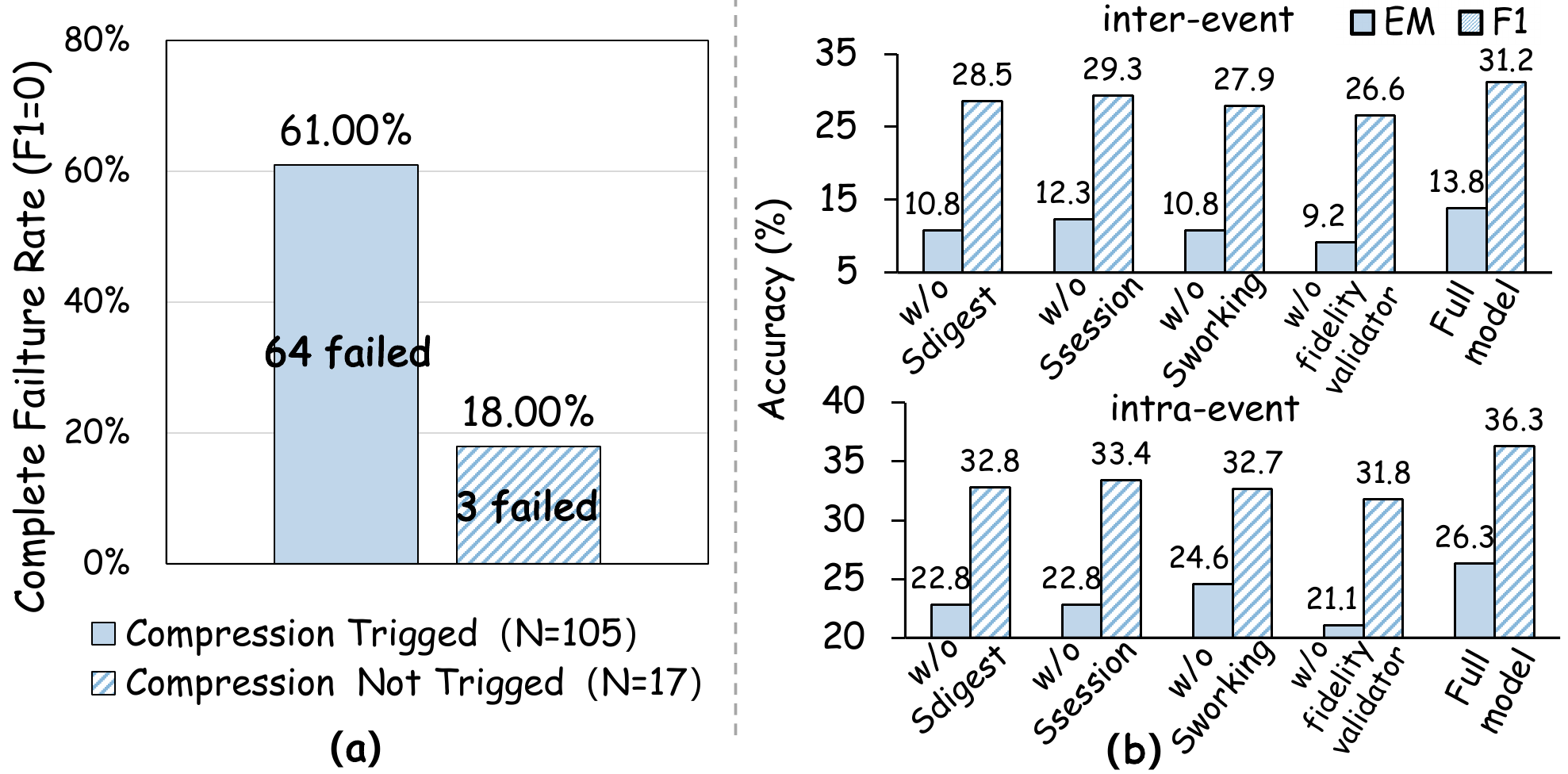}
    \caption{(a) Comparison of failure rates between cases with and without unconstrained compression. (b) Ablation of components in working Memory.}
    \label{fig:ab_working_memory}
    \vspace{-4mm}
\end{figure}

\noindent \textbf{Effectiveness of working memory.} Figure~\ref{fig:ab_working_memory}(a) shows that compression is frequently triggered during task execution, yet unconstrained compression causes errors in up to 61\% of cases, indicating that naive compression fails to preserve reliable reasoning states. Consistently, removing working memory leads to a 4.7 drop in Table~\ref{tab:ablation_memory}, highlighting the importance of maintaining valid intermediate states. We further evaluate its internal summaries and fidelity mechanism (Figure~\ref{fig:ab_working_memory}(b)). Ablating any summary component degrades performance, indicating that step-level traces, session-level decisions, and current plans jointly support traceability, strategic continuity, and task continuation. The fidelity mechanism further preserves key metadata during compression, yielding additional gains.

\noindent \textbf{Effectiveness of episodic memory.} Omitting episodic memory leads to a 4.4 drop (Table~\ref{tab:ablation_memory}), indicating that agents without explicit post-hoc verification struggle to reconstruct spatiotemporal reasoning trajectories during long horizon search. Once errors arise in intermediate steps, they can accumulate and result in execution drift, as further shown in Figure~\ref{fig:intro2}. Accordingly, episodic memory mitigates this issue by performing reflective self-correction across multiple audit dimensions after a candidate answer is generated, thereby improving logical consistency in complex multi hop queries.

\begin{figure}[!t]
    \centering
    \includegraphics[width=\linewidth]{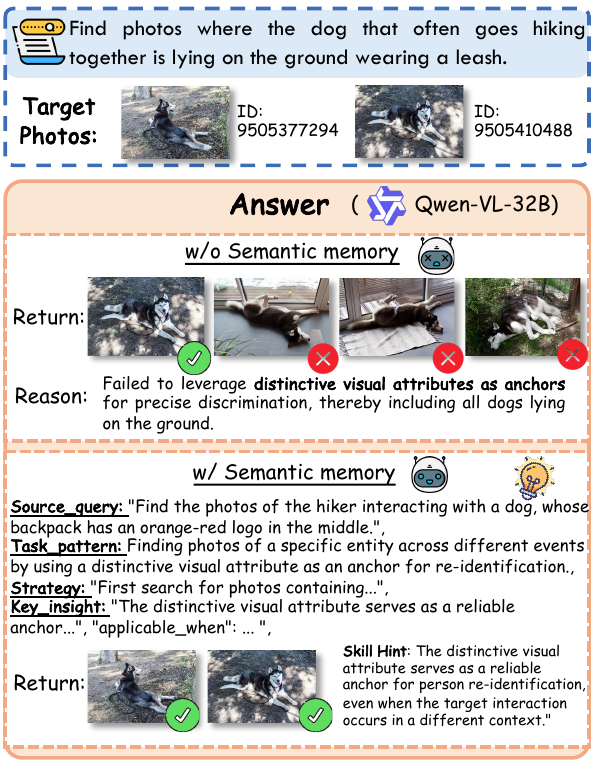}
    \caption{Example of how agents complete the same task with and without semantic memory.}
    \label{sem_example}
    \vspace{-3.5mm}
\end{figure}

\noindent \textbf{Effectiveness of semantic memory.} Removing semantic memory results in a 4.0 drop (Table~\ref{tab:ablation_memory}), showing that prior reasoning trajectories provide important guidance for subsequent reasoning. Figure~\ref{sem_example} provides an intuitive visualization of how semantic memory benefits the agent. Specifically, semantic memory decontextualizes successful trajectories into reusable knowledge, allowing the agent to exploit prior experience to improve reasoning and support more flexible cross-task orchestration. We further analyze the design rationale of this skill distillation strategy in Section~\ref{More Analysis}.

\subsection{More Analysis}
\label{More Analysis}

\noindent \textbf{Is direct retrieval sufficient for Deep Image Search?} We evaluate Qwen3-VL-Embedding, including its 2B and 8B variants, as representative leading vision-language embedding models. As shown in Table~\ref{tab:direct_retrieval}, both models perform poorly under direct retrieval on DISBench, with MAP@3 around 10\% and Recall@3 ranging from 10.1\% to 12.5\%. These results indicate that the bottleneck stems from the retrieval paradigm rather than model capacity. By independently scoring each image, conventional retrieval~\citep{traditionalretrieval} fails to capture cross-image associations and context-dependent constraints, highlighting the need for multi-step reasoning over corpus-level context.

\begin{table}[!t]
\centering
\setlength{\tabcolsep}{2pt}
\renewcommand{\arraystretch}{1.2}
\small
\caption{Direct retrieval performance of Qwen3-VL-Embedding models on DISBench.}
\label{tab:direct_retrieval}
\begin{tabularx}{\columnwidth}{@{}ccccccc@{}}
\toprule
\multirow{2}{*}{\centering\textbf{Direct retrieval Model}} 
& \multicolumn{3}{c}{\textbf{MAP}} 
& \multicolumn{3}{c}{\textbf{Recall}} \\
\cmidrule(lr){2-4} \cmidrule(lr){5-7}
& \textbf{@1} & \textbf{@3} & \textbf{@5} 
& \textbf{@1} & \textbf{@3} & \textbf{@5} \\
\midrule
Qwen3-VL-Embedding-2B 
& 11.8 & 9.9 & 10.3 
& 3.6 & 10.1 & 14.1 \\
Qwen3-VL-Embedding-8B 
& \textbf{13.1} & \textbf{11.4} & \textbf{12.4} 
& \textbf{4.8} & \textbf{12.5} & \textbf{17.8} \\
\bottomrule
\end{tabularx}
\end{table}

\begin{figure}[!t]
    \centering
    \includegraphics[width=\linewidth]{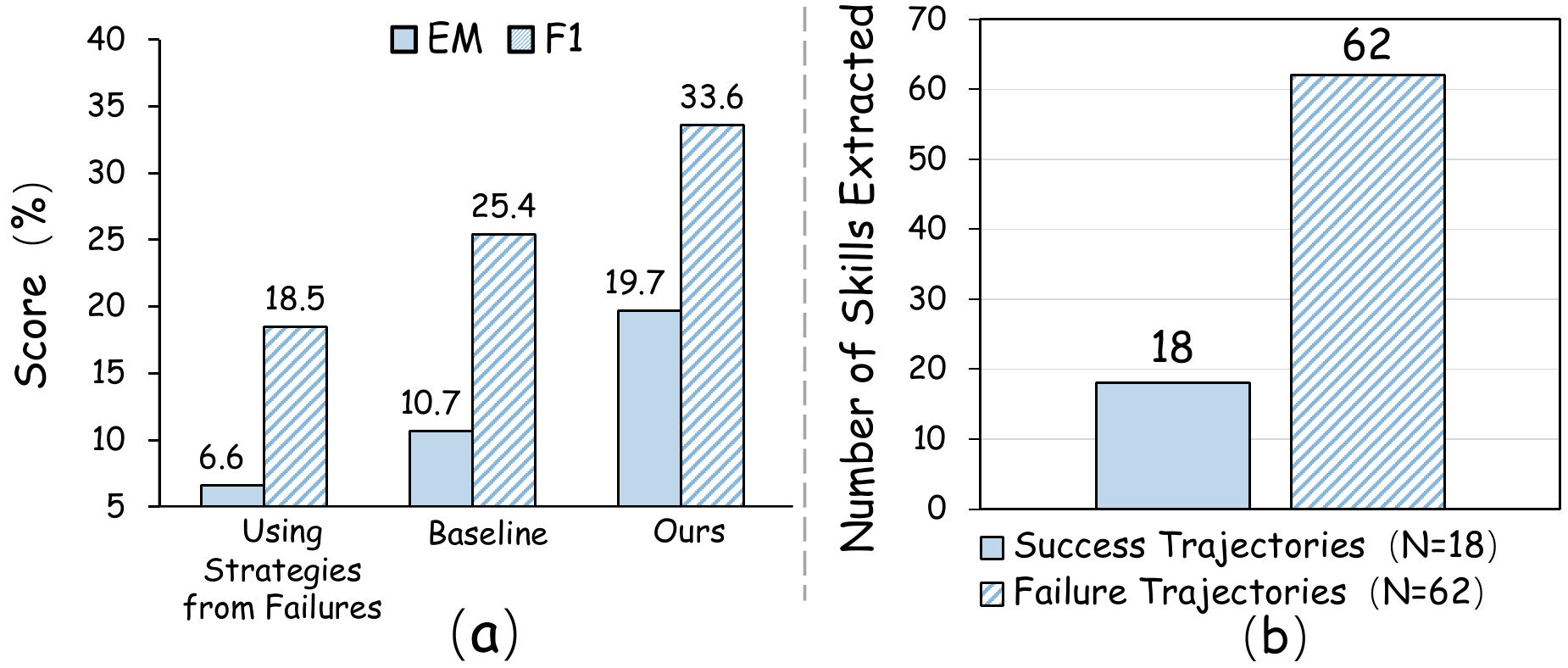}
    \caption{(a) Comparison among different skill distillation strategies. (b) Comparison of the number of skills extracted from successful and failure trajectories.}
    \label{fig:ab_skill}
    \vspace{-5mm}
\end{figure}


\noindent \textbf{Why distill skills from successes only?} Following prior work~\citep{xskill,reme}, we further examine whether lessons extracted from failure trajectories can complement skill distillation from successful trajectories. As shown in Figure~\ref{fig:ab_skill}(a), using skills distilled from failure trajectories does not improve performance and instead leads to a substantial decline. We attribute this result to the sparse yet complex nature of DISBench~\citep{deepimagesearch}. Although the benchmark contains relatively few queries, each query typically involves multi-step reasoning and diverse tool calls, resulting in highly heterogeneous failure causes and corresponding remedies, as shown in Figure~\ref{fig:ab_skill}(b). In contrast, successful trajectories more readily converge into compact and reusable strategies, whereas failure trajectories tend to yield vague and weakly actionable heuristics.

\begin{figure}[!t]
    \centering
    \includegraphics[width=\linewidth]{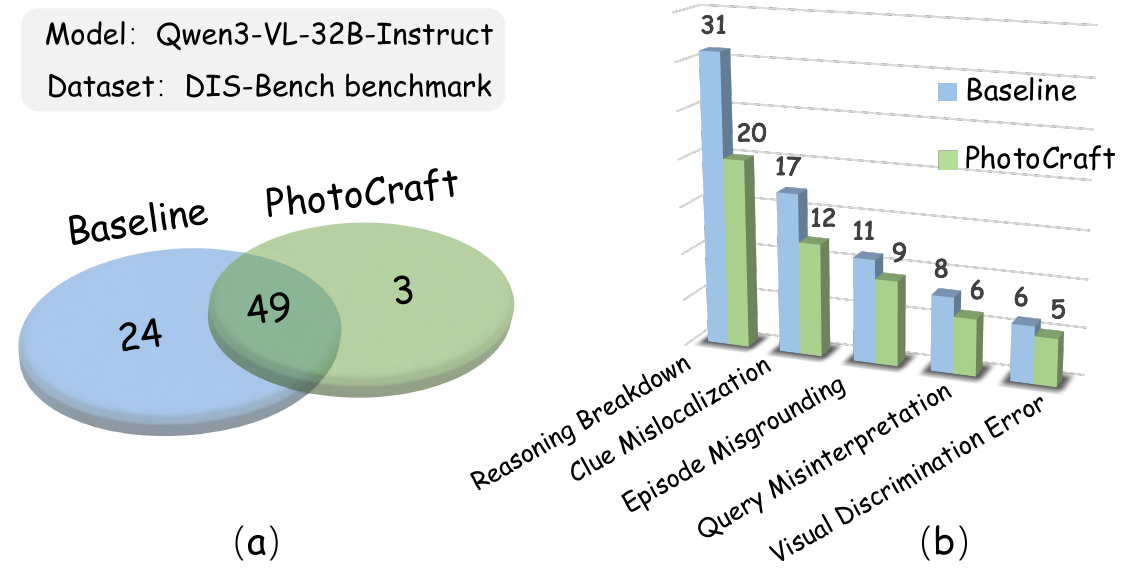}
    \vspace{-7mm}
    \caption{Error analysis with and without PhotoCraft.}
    \label{fig:error_fenxi}
    \vspace{-3mm}
\end{figure}

\begin{figure}[!t]
    \centering
    \includegraphics[width=\linewidth]{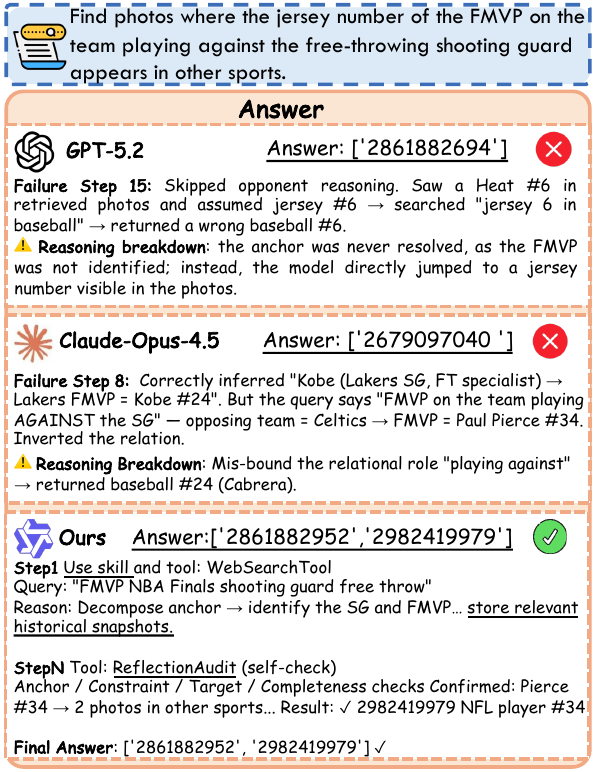}
    \vspace{-7mm}
    \caption{Visualization of responses and reasoning by different methods for a specific question.}
    \label{reasoning_example}
    \vspace{-3.5mm}
\end{figure}

\noindent \textbf{Error Analysis.} We analyze error patterns with and without PhotoCraft. The Venn diagram (Figure~\ref{fig:error_fenxi}(a)) shows that the total number of failure cases changes from 73 in the baseline to 52 with PhotoCraft. Notably, PhotoCraft corrects \textbf{18} baseline specific errors and introduces 3 new errors. We further manually review and categorize each failure case to examine the impact of PhotoCraft on different error types, as shown in Figure~\ref{fig:error_fenxi}(b). All error types exhibit consistent improvement, with Reasoning Breakdown decreasing from \textbf{31} to \textbf{20}. Figure~\ref{reasoning_example} further shows that ImageSeeker fails under both GPT-5.2 and Opus-4.5 due to reasoning breakdown. In contrast, PhotoCraft successfully resolves the query. These results indicates that PhotoCraft can effectively leverage its memory system to strengthen multi step reasoning, reduce the propagation of earlier errors, and improve generalization through its self evolving mechanism.

\section{Conclusion}
\label{Conclusion}

To address the difficulty of memoryless agents in maintaining long-horizon context and transferring experience across tasks in Deep Image Search, we propose PhotoCraft, a training-free, hierarchical memory framework that equips photo-search agents with working, episodic, and semantic memory. These memory modules respectively support intermediate-state maintenance, reflective self-correction, and reuse of successful strategies, thereby mitigating execution drift and experience isolation. By continuously updating and retrieving memory during reasoning, PhotoCraft enables agents to perform more stable multi-step contextual reasoning over complex visual histories and more flexible cross-task orchestration. Extensive experiments demonstrate its effectiveness for reliable and generalizable Deep Image Search.

\section{Limitations}
\label{Limitations}
This paper focuses on hierarchical self-evolving memory for Deep Image Search over personal photo collections. Although PhotoCraft demonstrates consistent improvements in multi-step reasoning and cross-task experience transfer, several limitations remain to be addressed in future work.

\noindent \textbf{Benchmark Scale.} PhotoCraft is evaluated only on DISBench~\citep{deepimagesearch}, as large-scale real-world personal photo benchmarks entail substantial privacy risks due to sensitive album content, timestamps, locations, and cross-event associations. DISBench’s rigorous human filtering ensures data quality, task reliability, and reproducibility, but its 6.1\% retention rate also leads to a limited query scale. Future work should develop privacy-preserving data construction and anonymization protocols to expand coverage while maintaining annotation quality.

\noindent \textbf{Skill Coverage.} PhotoCraft’s semantic memory relies on transferable skills distilled from successful historical trajectories. While this mechanism mitigates redundant planning and experience isolation, its coverage is still bounded by the diversity of observed tasks and successful episodes. Future work may explore finer-grained skill abstraction, hierarchical skill organization, and active skill discovery to improve semantic memory for long-tail queries and complex compositional reasoning.


\section{Ethics Statement}
\label{Ethics Statement}

We acknowledge that retrieval over personal photo collections inevitably raises privacy concerns. This issue is not only inherent to research on personal data understanding, but also broadly shared by the personalized systems community. To mitigate these risks, DISBench~\citep{deepimagesearch} is built on images from YFCC100M~\citep{yfcc100m} that were publicly shared under Creative Commons licenses. Given that this task requires mining cross-image and cross-event associations within personal photo collections and involves complex intent modeling and contextual retrieval demands, it is difficult for us to conduct additional evaluations on other real private data without sufficient authorization or on existing general-purpose retrieval datasets~\citep{mscoco,retref2,retref3}. We emphasize that the intended use of this technology is to help users retrieve, organize, and revisit their own visual memories, rather than to analyze others’ data without consent.



\bibliography{custom}

\appendix

\section{Appendix}
\label{sec:appendix}

\subsection{Details of DISBench}
\label{Details of DISBench}

\begin{figure}[!t]
    \centering
    \includegraphics[width=\linewidth]{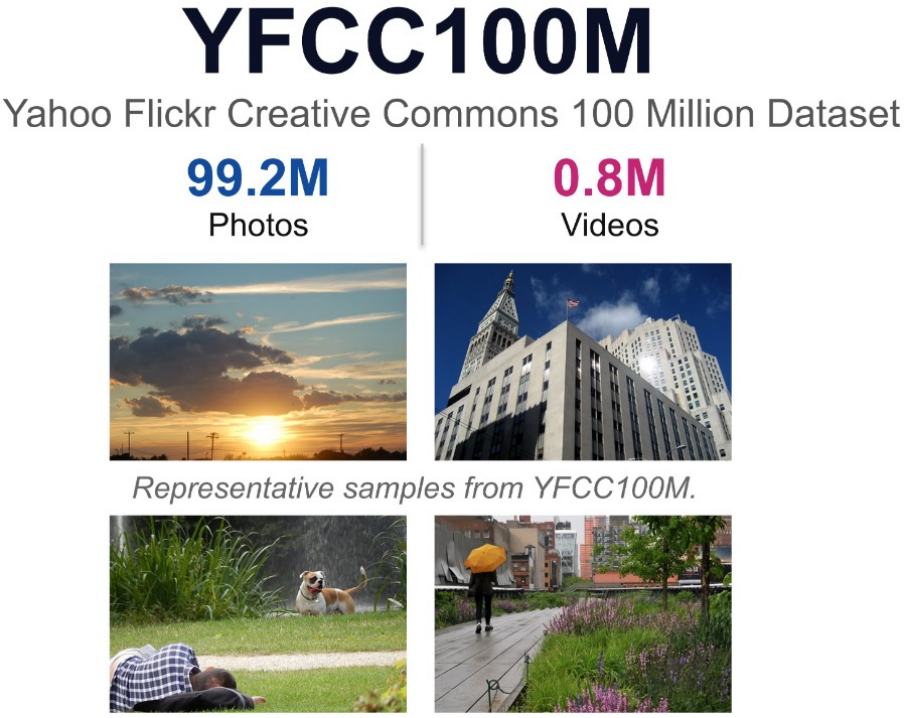}
    \caption{Overview of YFCC100M as the data source of DISBench. The dataset provides large-scale photos and videos with temporal, geographic, tag, and media metadata, supporting context-grounded deep image search.}
    \label{fig:dataset}
    \vspace{-5mm}
\end{figure}

\noindent \textbf{Source dataset---YFCC100M.} DISBench~\citep{deepimagesearch} is derived from YFCC100M~\citep{yfcc100m}, a Creative Commons Flickr collection with roughly 99.2M photos and 0.8M videos, together with timestamps, geolocations, tags, and media metadata. This source fits PhotoCraft because Deep Image Search requires visual evidence grounded in personal-history context, as shown in Figure~\ref{fig:dataset}. Although Flickr sharing patterns are not demographically exhaustive, the open license, stable metadata schema, and long-standing multimedia usage make the benchmark reproducible and transferable to other temporally or geographically annotated photo collections.

\begin{figure*}[!t]
    \centering
    \includegraphics[width=\textwidth]{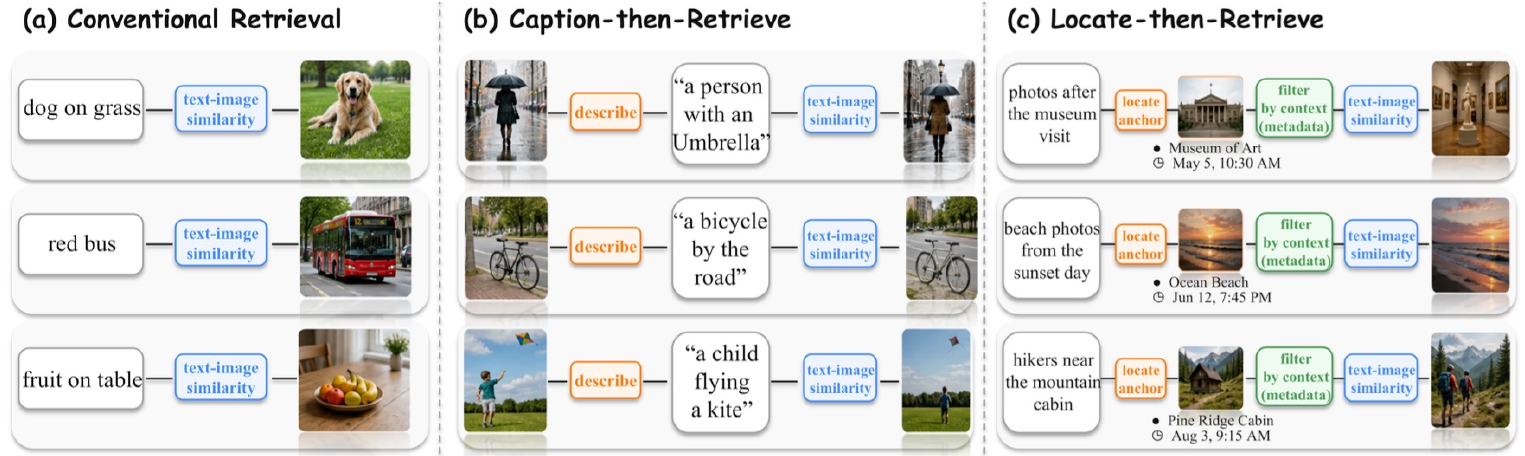}
    \caption{Comparison of image retrieval paradigms. Unlike conventional retrieval and caption-then-retrieve methods, PhotoCraft first localizes contextual anchors and then performs metadata-aware retrieval to support deep image search.}
    \label{fig:task}
\end{figure*}

\noindent \textbf{Task Description.}
DISBench instantiates Deep Image Search as a context-aware set retrieval task over personal visual histories. Given a photo corpus with visual content and associated metadata, such as timestamps and geolocations, the model is required to return all images that satisfy a natural-language query. As illustrated in Figure~\ref{fig:task}, this setting differs fundamentally from conventional retrieval paradigms. Conventional retrieval directly matches a query to each image through text-image similarity~\citep{imageret}, while caption-then-retrieve methods first convert images into textual descriptions but still perform largely independent instance-level matching~\citep{composed2,comsurvey}. In contrast, DISBench requires a locate-then-retrieve process. The target images are often not identifiable from their visual appearance alone; instead, the model must first localize contextual anchors, such as relevant events, places, time periods, or visually grounded clues, and then perform metadata-aware retrieval within the localized context. This design makes the task depend on corpus-level contextual reasoning rather than isolated semantic matching. Therefore, DISBench evaluates whether models can integrate visual evidence, temporal and geographic metadata, and cross-image associations to construct evidence chains and retrieve the complete target set, which requires an effective memory system to support multi-step contextual reasoning.

\subsection{Agent Implementation Details}
\label{Agent Implementation Details}

\begin{table*}[!t]
\centering
\setlength{\tabcolsep}{3pt}
\renewcommand{\arraystretch}{1.15}
\small
\caption{Tool interfaces used by PhotoCraft. The tools support visual retrieval, metadata access, candidate filtering, photo inspection, web lookup, and memory condensation during multi-step photo search.}
\label{tab:tool_descriptions}
\begin{tabularx}{\textwidth}{@{}>{\raggedright\arraybackslash}p{0.18\textwidth}>{\raggedright\arraybackslash}p{0.38\textwidth}>{\raggedright\arraybackslash}X@{}}
\toprule
\textbf{Tool} & \textbf{Role in PhotoCraft} & \textbf{Usage} \\
\midrule
\textbf{ImageSearch} &
Retrieves candidate photos from the visual history using textual and optional visual cues. PhotoCraft uses it to locate event anchors, collect target candidates, and revisit the corpus when memory audit identifies missing evidence. &
The controller issues short search queries, optionally scopes the search to a saved candidate set, and receives ranked photo IDs with retrieval scores for later inspection. \\
\midrule
\textbf{GetMetadata} &
Obtains structured metadata for selected photos. It grounds visual hypotheses in time and location evidence, which is critical for queries involving events, trips, or temporal relations. &
Given photo IDs, it returns fields such as timestamp and address; PhotoCraft preserves these fields in working memory to avoid lossy compression of task constraints. \\
\midrule
\textbf{FilterMetadata} &
Filters the visual history according to inferred temporal or geographic constraints. It converts an identified anchor event into a deterministic candidate pool. &
The controller supplies a metadata expression and receives matching photo IDs, which may be used as a scoped search space for subsequent retrieval or verification. \\
\midrule
\textbf{ViewPhotos} &
Loads selected photos into the multimodal context for direct visual inspection. It is used to verify anchors, distinguish visually similar candidates, and confirm final targets. &
Given a bounded list of photo IDs, it returns the corresponding images with explicit identifiers so that visual evidence can be linked to the final answer. \\
\midrule
\textbf{WebSearch} &
Provides auxiliary external knowledge for entities that cannot be resolved from the album alone, such as landmarks, venues, or event names. &
The controller issues a web query and uses returned snippets only as supporting context; final decisions must still be grounded in album evidence. \\
\midrule
\textbf{Memory Condense} &
Maintains a compact reasoning state when the interaction becomes long. It does not access new photos, but activates the working-memory mechanism described in Section~\ref{Hierarchical Self-Evolving Memory}. &
It compresses older interaction history into message digest, session memory, and working memory while retaining metadata-critical evidence such as photo IDs, dates, and locations. \\
\bottomrule
\label{Tool interfaces used by PhotoCraft}
\end{tabularx}
\end{table*}

\subsubsection{Tool Description}
\label{Tool Description}

PhotoCraft is implemented as a tool-augmented photo-search agent. The tool interface provides access to visual retrieval, metadata lookup, metadata-based filtering, photo inspection, and auxiliary web knowledge, detailed descriptions of the specific tools are provided in Table~\ref{Tool interfaces used by PhotoCraft}. During each interaction step, the controller selects a tool, observes its output, and records task-relevant evidence in memory for subsequent search, verification, or answer generation. The controller composes these tools according to the query and the current memory state. Working memory keeps active constraints and intermediate evidence available during search, episodic memory audits whether the accumulated trace supports the proposed answer, and semantic memory provides reusable strategies from previous episodes. Together, the tool interface and memory modules support iterative retrieval, grounding, and verification over long visual histories.

\subsubsection{Experimental Configurations}
\label{Experimental Configurations}

\textbf{Models.} For agentic evaluation, we access proprietary models through their official APIs, while open-source models are served locally with vLLM via an OpenAI-compatible API endpoint. Following ImageSeeker~\citep{deepimagesearch}, all models use their default temperature settings and are equipped with the same tool interfaces. For the ImageSearch tool, we adopt Qwen3-VL-Embedding~\citep{qwen3} as the multimodal encoder. We evaluate two encoder scales, 2B and 8B parameters, to examine how retrieval quality affects overall agent performance.

\begin{figure}[!t]
    \centering
    \includegraphics[width=\linewidth]{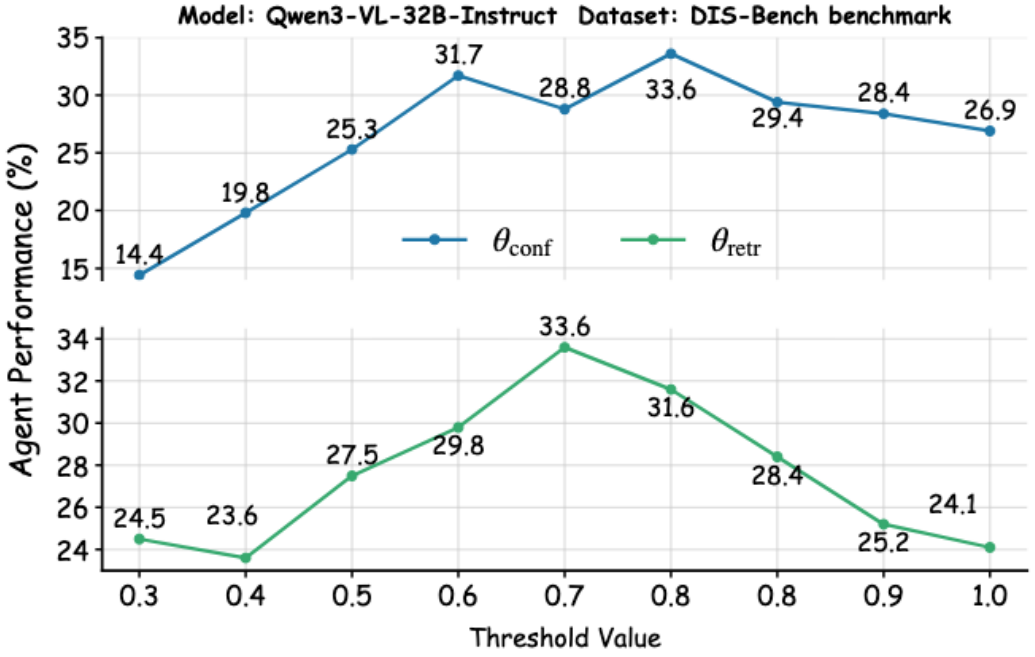}
    \caption{Effect of the skill-distillation confidence threshold $\theta_{\mathrm{conf}}$ and retrieval similarity threshold $\theta_{\mathrm{retr}}$ on agent performance (\%) on DISbench.}
    \label{fig:threshold}
\end{figure}

\noindent \textbf{Hyperparameters.} Consistent with ImageSeeker’s setting~\citep{deepimagesearch}, we set the maximum number of interaction turns to 30 and require the agent to return an answer once this budget is exhausted. 
Context compression is performed by GPT-4o~\citep{gpt4o}, and ImageSearch returns 20 results by default for each retrieval operation. 
To select appropriate thresholds, we analyze the relationship between performance and two key hyperparameters: the skill-distillation confidence threshold $\theta_{\mathrm{conf}}$ and the retrieval similarity threshold $\theta_{\mathrm{retr}}$, sweeping both from 0.3 to 1.0. 
As shown in Figure~\ref{fig:threshold}, increasing $\theta_{\mathrm{conf}}$ initially improves performance by filtering unreliable trajectories and improving the quality of distilled skills. 
However, overly large values reduce the coverage of semantic memory, as fewer trajectories satisfy the distillation criterion. 
Similarly, moderate values of $\theta_{\mathrm{retr}}$ prevent weakly related skills from being injected, whereas excessive thresholds hinder the agent from reusing relevant prior experience. 
Based on these observations, we set $\theta_{\mathrm{conf}}$=0.85 and $\theta_{\mathrm{retr}}$=0.65 in the main experiments.

\subsection{Details of Prompts}
\label{Details of Prompts}

\subsubsection{Prompts for Episodic Audit Signals}
\label{Prompts for Episodic Audit Signals}

\begin{tcolorbox}[
  title={Reasoning Chain Audit Prompt},
  sharp corners,
  breakable,
  colframe=auditgreen,
  colback=auditlightgreen,
  boxrule=2pt,
  boxsep=0.5pt,
  left=2pt,
  right=2pt,
  top=2pt,
  bottom=2pt,
  enhanced,
  shadow={3pt}{-3pt}{0pt}{opacity=0.8,auditgray}
]
\begin{lstlisting}[style=auditpromptstyle]
reflection_prompt = (
    "[REFLECTION: Reasoning Chain Audit]\n"
    f"Your proposed answer: {pred_ids}\n\n"
    "Before finalizing, perform these checks:\n"
    "1. ANCHOR CHECK: Did you correctly identify the anchor event? "
    "Could a different event/entity be the correct anchor? "
    "Did you visually verify it?\n"
    "2. CONSTRAINT CHECK: Did you derive the correct date/location "
    "from the anchor? Did you use a date RANGE (not exact timestamp)?\n"
    "3. TARGET CHECK: For each proposed photo, does it satisfy the "
    "TARGET criteria specifically? Remove any that only match the "
    "anchor but not the target.\n"
    "4. COMPLETENESS: Are there additional matching photos you missed? "
    "Consider trying different search keywords.\n"
    "5. DECISION: Remove non-matching photos, add newly discovered ones, "
    f"then give your revised final answer.{extra_notes}\n"
)
\end{lstlisting}
\end{tcolorbox}

\subsubsection{Prompts for Semantic Skill Entries}
\label{Prompts for Semantic Skill Entries}

\begin{tcolorbox}[
  title={Strategy Extraction Prompt},
  sharp corners,
  breakable,
  colframe=extractionorange,
  colback=extractionlightorange,
  boxrule=2pt,
  boxsep=0.5pt,
  left=2pt,
  right=2pt,
  top=2pt,
  bottom=2pt,
  enhanced,
  shadow={3pt}{-3pt}{0pt}{opacity=0.8,fill=extractiongray,draw=extractiongray}
]
\begin{lstlisting}[style=extractionpromptstyle]
EXTRACTION_PROMPT = """\
You are analyzing a successful photo-search trajectory to extract a reusable strategy.

## Tools available to the agent
- ImageSearchTool: keyword-based visual search across the photo collection
- ViewPhotosTool: view specific photos by ID
- GetMetadataTool: get date/location metadata of photos
- FilterMetadataTool: filter photos by date/location expressions
- WebSearchTool: web lookup for external facts

## Successful Query
Query: {query}

## Agent Trajectory (abbreviated)
{trajectory}

## Your Task
Extract ONE concise, reusable search strategy from this successful execution.
Output **only** a JSON object (no other text):

{{
  "task_pattern": "<unique natural-language description of the SPECIFIC query structure -- MUST reflect what makes THIS query different from other types>",
  "strategy": "<2-3 sentences: step-by-step tool-usage and reasoning approach that worked. NO entity names or photo IDs.>",
  "key_insight": "<1 sentence: the single most important takeaway>",
  "applicable_when": "<1 sentence: describe PRECISELY when this strategy helps -- be SPECIFIC about the query structure, not generic>"
}}

CRITICAL RULES:
- NO photo IDs, user IDs, person names, place names, or specific entities
- The task_pattern MUST be SPECIFIC to this query's structure. Different queries need different patterns.
- Do NOT default to generic descriptions -- if the agent solved it via direct keyword search and visual verification, say that
- The strategy must help an agent facing a SIMILAR (not identical) query
- Be CONCISE and ACTIONABLE
"""
\end{lstlisting}
\end{tcolorbox}

\begin{figure*}[!t]
\centering
\begin{tcolorbox}[
  title={System Prompt},
  width=\textwidth,
  sharp corners,
  colframe=promptblue,
  colback=white,
  boxrule=1.2pt,
  boxsep=1pt,
  left=2pt,
  right=2pt,
  top=2pt,
  bottom=2pt,
  enhanced,
  shadow={2pt}{-2pt}{0pt}{opacity=0.6,promptgray}
]
\begin{lstlisting}[style=promptstyle]
SYSTEM_PROMPT = \
"""You are the Deep Image Search Agent. Your job is to find specific photos in a user's personal photo collection through multi-step reasoning over images and metadata.

YOUR TASK

1. QUERY ANALYSIS -- decompose the query BEFORE calling any tool
   A. ANCHOR: Identify the reference event, entity, or photo mentioned in the query (e.g., "the concert identified by the blue logo", "the day when you saw the nude statue"). This is what you search for FIRST to derive constraints.
   B. CONSTRAINT: What fact will the anchor give you? (a specific date, location, person identity, or visual attribute)
   C. TARGET: What the final answer photos must depict. Carefully separate anchor criteria from target criteria -- anchors help you FIND the targets but need not appear IN the target photos.
   D. Write out your ANCHOR -> CONSTRAINT -> TARGET decomposition and your step-by-step plan before calling any tool.

2. EXECUTE WITH STEP REFLECTION -- after EACH tool result, assess before your next action:
   - What did I just learn? Does it confirm or change my plan?
   - Am I still pursuing the right subgoal, or have I drifted?
   - If results are poor, I should try DIFFERENT keywords or a different approach -- NOT repeat the same failing strategy.
   - After finding the anchor, I must extract its date/location IMMEDIATELY and use it to constrain the next search.

3. VERIFY BEFORE ANSWERING
   - For each candidate photo, confirm it satisfies the TARGET criteria (visual content + metadata constraints).
   - Anchor constraints define the event/context; do NOT require anchor visuals in target photos unless the query explicitly says so.
   - When a query uses temporal phrases ("on the day we...", "during the trip to..."), treat this as a time/location constraint, not a visual requirement.
   - Return ALL photos that meet the intent. If nothing perfectly matches, return closest candidates.

EXECUTION RULES

1. SEARCH PRECISION
   Use SHORT keywords (2-5 words) for ImageSearchTool, NOT the full query sentence.
   Good: "leopard print parade", "dinosaur skeleton museum"
   Bad:  "Find photos from the Royal Regiment of Scotland band parade that include leopard print clothing"
   For complex queries, run MULTIPLE searches with DIFFERENT short keywords.

2. ANCHOR FIRST
   For queries referencing other events ("on the day when...", "the person who...", "after visiting..."):
   - Find the ANCHOR event/photo first.
   - VISUALLY VERIFY the anchor with ViewPhotosTool (do not assume based on search score alone).
   - If uncertain, check MULTIPLE anchor candidates before committing.
   - Extract the date/location from anchor metadata, then search for targets within that constraint.

3. DATE FILTERING
   When filtering by date, ALWAYS use a date RANGE or prefix match:
   Correct: time >= '2013-06-09' and time < '2013-06-10'
   WRONG:   time == '2013-06-09 20:36:00'  (matches only ONE SECOND!)

4. MULTI-ATTEMPT SEARCH
   If your first search returns few or no plausible matches, try AT LEAST 2 DIFFERENT keyword variants.
   Example: "Sydney Harbour Bridge night" -> also try "bridge night cityscape" -> also try "harbour bridge evening"

5. SCOPE CONTROL
   After deriving a date/location from the anchor, constrain your target search to that scope.
   Do NOT search the entire album for targets that should be within a specific event.
   Use FilterMetadataTool or search_within to narrow the scope.

6. NEVER-EMPTY RULE
   You MUST NOT return an empty list []. If no perfect matches exist, return the closest candidates.

ANSWER FORMAT
In your final response, output exactly: "The final answer is: ["9876543210", "1234567890", ...]."
Always use ACTUAL photo IDs from your search results. NEVER use placeholder names like "photo_id1".
"""
\end{lstlisting}
\end{tcolorbox}
\label{fig:system_prompt}
\end{figure*}

\subsubsection{System Prompt}
\label{app:system_prompt}

\subsection{Qualitative Examples}
\label{Qualitative Examples}
To illustrate how PhotoCraft navigates visual histories, we present a representative successful reasoning trajectory in Figure~\ref{fig:reasoning-example}. By integrating semantic guidance, hierarchical memory updates, and tool-augmented verification, PhotoCraft constructs a coherent evidence chain that enables reliable multi-step contextual reasoning for Deep Image Search.

\onecolumn

\begin{tcolorbox}[
    title={Reasoning Example of PhotoCraft on Deep Image Search},
    width=\textwidth,
    sharp corners,
    breakable,
    enhanced jigsaw,
    colframe=gray,
    colback=white,
    boxrule=1.2pt,
    boxsep=1pt,
    left=4pt,
    right=4pt,
    top=4pt,
    bottom=4pt,
    shadow={2pt}{-2pt}{0pt}{opacity=0.6,promptgray}
]
\begin{lstlisting}[style=promptstyle, breaklines=true, escapeinside={(*@}{@*)}]
(*@\textbf{\texttt{User Query:}}@*)
"Identify the date of the photo showing a person holding a glass item first manufactured in 1753. Then find all photos of seaside sunsets taken during the next calendar month after that date."

(*@\textbf{\texttt{Ground Truth:}}@*)
5 photos: [5981883283, 5981890977, 5978559064, 5978560626, 5982435276]

(*@\textbf{\texttt{Retrieved Skill:}}@*)
Past-experience hint: "Visual similarity search within a temporally and spatially constrained subset is more reliable than searching the entire collection."

(*@\textbf{\texttt{Query Decomposition:}}@*)
Anchor: A photo showing a person holding the glass item indicated by the historical clue.
Constraint: The shooting date of the anchor photo, from which the next calendar month must be derived.
Target: All photos depicting seaside sunsets taken during that next calendar month.
Key distinction: The anchor photo is only used to infer the temporal constraint; the final target photos do not need to contain the glass item.

(*@\textbf{\texttt{Reasoning Process:}}@*)

(*@\par\Needspace{12\baselineskip}@*)
(*@\underline{\texttt{Step 1: Historical Fact Resolution}}@*)
Tool: WebSearchTool
Query: "glass item first manufactured in 1753"
Reason:
    The query contains an implicit historical clue. Before searching the image collection, the agent resolves the clue into a concrete visual concept.
Result:
    The glass item is resolved as a glass bottle.
Inference:
    The anchor image should likely contain a person holding a glass bottle.

(*@\par\Needspace{12\baselineskip}@*)
(*@\underline{\texttt{Step 2: Anchor Candidate Retrieval}}@*)
Tool: ImageSearchTool
Query: "person holding glass bottle"
Reason:
    Use concise visual keywords to localize candidate anchor photos.
Result:
    Top candidates:
        5890572248, similarity = 0.30
        6782065157, similarity = 0.22
        7662314970, similarity = 0.21
        6120773892, similarity = 0.19
        5680471152, similarity = 0.19
        6789371936, similarity = 0.19
        6782087367, similarity = 0.18
        7662302726, similarity = 0.16
        6667515055, similarity = 0.16
        5993266842, similarity = 0.15
Observation:
    The top-ranked candidate 5890572248 has the highest similarity and is separated from the remaining candidates by a clear score gap.

(*@\par\Needspace{12\baselineskip}@*)
(*@\underline{\texttt{Step 3: Anchor Candidate Inspection}}@*)
Tool: ViewPhotosTool
Photos:
    [5890572248, 6782065157, 7662314970, 6120773892, 5680471152]
Reason:
    The anchor must be visually verified instead of selected by score alone.
Inspection:
    5890572248:
        Shows a person clearly holding a glass bottle.
        Matches both the human-object interaction and the resolved glass-item clue.
    6782065157 / 7662314970 / 6120773892 / 5680471152:
        Visually weaker candidates for the anchor condition.
        They do not match the "person holding glass bottle" description as directly as 5890572248.
Decision:
    Select 5890572248 as the anchor image.

(*@\par\Needspace{12\baselineskip}@*)
(*@\underline{\texttt{Step 4: Anchor Metadata Extraction}}@*)
Tool: GetMetadataTool
Arguments:
    photos = [5890572248]
    fields = [time]
Reason:
    The query asks for the date of the anchor photo, which determines the target time window.
Result:
    5890572248:
        time = 2011-06-17 18:21:07
Inference:
    The anchor date is 2011-06-17.
    The next calendar month is July 2011.
    Therefore, the target time range is:
        [2011-07-01, 2011-08-01)

(*@\par\Needspace{14\baselineskip}@*)
(*@\underline{\texttt{Step 5: Temporal Candidate Subset Construction}}@*)
Tool: FilterMetadataTool
Arguments:
    time in [2011-07-01, 2011-08-01)
    save_as = "july_2011_photos"
Reason:
    Apply the retrieved skill by first constructing a temporally constrained candidate subset.
Result:
    257 photos are saved into the named subset "july_2011_photos".
Intermediate inspection:
    The first retrieved July photos include landscape, travel, and non-sunset scenes.
    This confirms that metadata filtering alone is insufficient; visual filtering is still required.

(*@\par\Needspace{14\baselineskip}@*)
(*@\underline{\texttt{Step 6: Constrained Visual Search for Target Concept}}@*)
Tool: ImageSearchTool
Arguments:
    text = "seaside sunset"
    search_within = "july_2011_photos"
Reason:
    Search only within the July 2011 subset to avoid visually similar sunsets from irrelevant months.
Result:
    Ranked candidates:
        5981883283, similarity = 0.40
        5981890977, similarity = 0.37
        5978559064, similarity = 0.37
        5978560626, similarity = 0.36
        5982435276, similarity = 0.36
        5974529368, similarity = 0.26
        5974534940, similarity = 0.23
        5977913213, similarity = 0.21
        5977896991, similarity = 0.19
        5973965359, similarity = 0.19
        5978452238, similarity = 0.19
        5978547224, similarity = 0.19
        5977996389, similarity = 0.18
        5982160580, similarity = 0.18
        5974531548, similarity = 0.18
Observation:
    The top-5 candidates form a compact high-confidence group with scores from 0.36 to 0.40.
    The similarity drops sharply from 0.36 at rank 5 to 0.26 at rank 6, suggesting a natural cutoff.

(*@\par\Needspace{14\baselineskip}@*)
(*@\underline{\texttt{Step 7: Visual Verification of Retrieved Targets}}@*)
Tool: ViewPhotosTool
Photos:
    [5981883283, 5981890977, 5978559064, 5978560626, 5982435276,
     5974529368, 5974534940, 5977913213]
Reason:
    Verify that the high-ranked photos depict seaside sunsets, and inspect lower-ranked candidates near the cutoff.
Verification:
    5981883283:
        Seaside scene with sunset colors over the ocean.
    5981890977:
        Seaside sunset with horizon and warm evening sky.
    5978559064:
        Ocean-side sunset scene with strong red-orange sky.
    5978560626:
        Seaside sunset with silhouetted foreground and ocean horizon.
    5982435276:
        Sunset over the sea with coastal elements.
Rejected near-cutoff candidates:
    5974529368 / 5974534940 / 5977913213:
        Related to coastal or outdoor scenes but do not satisfy the seaside-sunset target as clearly as the top-5.
Decision:
    Keep the top-5 candidates and reject lower-ranked near-cutoff images.

(*@\par\Needspace{14\baselineskip}@*)
(*@\underline{\texttt{Step 8: Metadata Verification of Target Photos}}@*)
Tool: GetMetadataTool
Arguments:
    photos = [5981883283, 5981890977, 5978559064, 5978560626, 5982435276]
    fields = [time]
Reason:
    Confirm that every visually valid target also satisfies the July 2011 temporal constraint.
Result:
    5981883283:
        time = 2011-07-24 17:04:09
    5981890977:
        time = 2011-07-24 17:05:30
    5978559064:
        time = 2011-07-22 17:29:46
    5978560626:
        time = 2011-07-22 17:30:37
    5982435276:
        time = 2011-07-24 17:03:38
Conclusion:
    All five candidates were taken within July 2011.

(*@\par\Needspace{14\baselineskip}@*)
(*@\underline{\texttt{Step 9: Completeness Verification with Alternative Query}}@*)
Tool: ImageSearchTool
Arguments:
    text = "ocean sunset"
    search_within = "july_2011_photos"
Reason:
    Use a semantically equivalent query to check whether additional valid seaside sunset photos were missed.
Result:
    Top candidates:
        5978559064, similarity = 0.38
        5982435276, similarity = 0.38
        5978560626, similarity = 0.36
        5981890977, similarity = 0.36
        5981883283, similarity = 0.34
        5974534940, similarity = 0.22
        5978547224, similarity = 0.22
        5974529368, similarity = 0.22
        5977956643, similarity = 0.20
        5977991819, similarity = 0.19
Observation:
    The same five photos appear as the top results under the alternative query.
    Lower-ranked results remain substantially weaker and do not introduce new valid targets.
Conclusion:
    The retrieved target set is complete.

(*@\par\Needspace{14\baselineskip}@*)
(*@\underline{\texttt{Step 10: Reflective Audit}}@*)
Audit checklist:
    Historical clue resolved:
        yes, the visual anchor concept is glass bottle.
    Anchor localized:
        yes, 5890572248 is visually verified as a person holding a glass bottle.
    Temporal constraint derived:
        yes, anchor time is 2011-06-17, so the next calendar month is July 2011.
    Temporal filtering applied:
        yes, target search is constrained to [2011-07-01, 2011-08-01).
    Target concept verified:
        yes, all retained photos depict seaside sunsets.
    Metadata verified:
        yes, all retained photos were taken in July 2011.
    Completeness checked:
        yes, the alternative query "ocean sunset" retrieves the same top-5 target set.

Final Answer:
[5981883283, 5981890977, 5978559064, 5978560626, 5982435276]
\end{lstlisting}
\end{tcolorbox}

\vspace{-2mm}
\captionof{figure}{An example of a successful reasoning trajectory of PhotoCraft on DISbench.}
\label{fig:reasoning-example}

\end{document}